\documentclass[runningheads]{llncs}
\usepackage[T1]{fontenc}
\usepackage{graphicx,verbatim}
\usepackage{algorithm}
\usepackage{algpseudocode}
\usepackage{booktabs}
\usepackage[misc]{ifsym}

\newcommand{\equalcontrib}{\fnmsep$^*$}
\newcommand{\corrauth}{\fnmsep\textsuperscript{\scalebox{0.75}{\Letter}}}
\begin{document}
\titlerunning{Human-in-the-Loop VDSS with Bandit Learning}
\title{Human-in-the-Loop Multi-Agent Ventilator Decision Support with Contextual Bandit Preference Learning}
%
%
\author{
Sijia Li\inst{1}\equalcontrib \and
Xiaoyu Tan\inst{2}\equalcontrib \and
Qixing Wang\inst{3} \and
Weiyi Zhao\inst{1} \and
Chen Zhan\inst{1} \and
Teqi Hao\inst{1} \and
Xuemin Wang\inst{4} \and
Lei Gu\inst{5} \and
Roland Eils\inst{6,7} \and
Xihe Qiu\inst{1}\corrauth
}

\authorrunning{S. Li et al.}

\institute{
Shanghai University of Engineering Science, Shanghai, China\\
\email{qiuxihe1993@gmail.com}
\and
Tencent Youtu Lab, Tencent, China
\and
Department of Critical Care Medicine, Shanghai Tenth People's Hospital, Tongji University School of Medicine, Shanghai, China
\and
Department of Emergency and Critical Disease, Songjiang Hospital Affiliated to Shanghai Jiao Tong University School of Medicine, Shanghai, China
\and
Max Planck Institute for Heart and Lung Research, Bad Nauheim, Germany
\and
Fudan University, Shanghai, China
\and
BIH at Charit\'e -- Universit\"atsmedizin Berlin, Berlin, Germany
}
\maketitle
\begingroup
\renewcommand{\thefootnote}{}
\footnotetext{\textsuperscript{*}Equal contribution. \quad \Letter~Corresponding author.}
\endgroup

\begin{abstract}
Ventilator decision support requires sequential decisions that track evolving physiology and disease trajectories while respecting safety boundaries and clinician specific tuning styles. Rule based approaches rarely generalize personalization, and end to end reinforcement learning or single large language model systems remain difficult to control and audit. We propose the \textbf{V}entilator \textbf{D}ecision \textbf{S}upport \textbf{S}ystem (\textbf{VDSS}), a human in the loop multi agent framework that coordinates modular decision components through contract driven structured interfaces and produces traceable evidence for review. VDSS performs online preference adaptation with a contextual bandit, updating clinician specific preferences from the final accepted decision at each adjustment cycle and using them to guide subsequent recommendations. Structured rejection feedback triggers targeted replanning to reduce unproductive iterations and improve interaction stability. Retrospective ICU trajectory replay with expert review indicates higher recommendation acceptability and fewer interaction rounds to reach an acceptable plan, supporting clinically deployable human AI collaboration.
\end{abstract}
\section{Introduction}
Mechanical ventilation is a central life sustaining therapy in the intensive care unit. Its clinical value extends beyond maintaining oxygenation and ventilation, as effective care requires continuous and fine grained adjustment of settings to match a patient’s evolving physiology \cite{topol2019high,fan2017official}. Ventilator management is inherently dynamic. Bedside teams must repeatedly revise decisions as lung mechanics, gas exchange, hemodynamics, and sedation levels change, while balancing multiple clinical objectives in real time \cite{acute2000ventilation}. At the same time, high quality titration depends on frequent assessment and rapid response by respiratory therapists and experienced clinicians \cite{kollef2000effect}. This specialized workforce is persistently scarce in many health systems, and sustained workload together with shift handoffs makes it difficult to deliver consistently individualized management even within the same institution \cite{gray2021changes}. As patient volume rises and workflow pressure increases, a practical motivation for ventilator decision support is to embed expert practice in a scalable and reusable form that reduces cognitive burden and improves consistency \cite{murali2024clinical}.


Current practice is guided by protocols and accumulated experience, yet real ICU care presents persistent challenges for decision support \cite{goodfellow2024aarc}. Patient heterogeneity and nonstationary trajectories cause strategy benefits to vary across individuals and over time, limiting fixed rule reliability \cite{jouvet2011development,ranieri2012acute}. Bedside titration often requires rapid setting changes based on multiple imperfect signals, and limited time to check and document the rationale can make adjustments inconsistent across cycles \cite{shi2023nurses,henry2022human}. Clinical teams further differ in risk tolerance, tuning style, and local conventions, and without explicit modeling and adaptation, decision support is unlikely to achieve sustained trust or uptake \cite{jivraj2023use,von2024adjustments}. Beyond a recommended setting, clinicians need a transparent link between the recommendation, the patient state, prevailing practice, and team preferences to support efficient repeated use \cite{munoz2025artificial}. These factors motivate ventilator decision support that is workflow aligned, coherent, and interpretable, while able to adapt as clinical preferences evolve \cite{rosenbacke2024explainable,bion2010human}.

Motivated by these needs, we argue that in high risk clinical settings, pointwise model performance is necessary but not sufficient \cite{slutsky2013ventilator,liu2020reporting}. Successful deployment instead hinges on decision support whose behavior can be inspected at the bedside and refined over time without breaking clinical workflow. We therefore propose the Ventilator Decision Support System, a human in the loop multi agent framework that decomposes ventilator decision making into modular agents and regulates inter agent communication through contract driven structured interfaces, enabling coherent multi round collaboration with a clear, auditable reasoning trace \cite{wu2024autogen,dewes2025contract}. The system uses layered short term and long term memory to manage trajectory level information and within round context, improving recommendation consistency and interaction stability \cite{packer2023memgpt}. It further incorporates a contextual bandit for online preference adaptation, treating the clinician’s final decision at the end of each adjustment cycle as feedback to update an individualized preference representation and guide subsequent recommendations. We validate this framework through retrospective ICU trajectory replay and expert review, showing improved recommendation acceptability and fewer interaction rounds to reach an acceptable recommendation \cite{cox2019effects}.

\begin{enumerate}
  \item VDSS is introduced as a human in the loop multi agent ventilator decision support framework that enables verifiable and maintainable collaboration through modular decomposition and contract driven interfaces.
  \item Layered short term and long term memory is integrated with an auditable evidence trail so that multi round recommendations remain coherent and are accompanied by traceable state summaries and rationale for structured review.
  \item An online preference adaptation mechanism is developed based on a contextual bandit that uses cycle level clinician decisions as feedback to adapt to different tuning styles, improving acceptability and reducing interaction rounds.
\end{enumerate}

\begin{figure}[t]
  \centering
  \includegraphics[width=\linewidth]{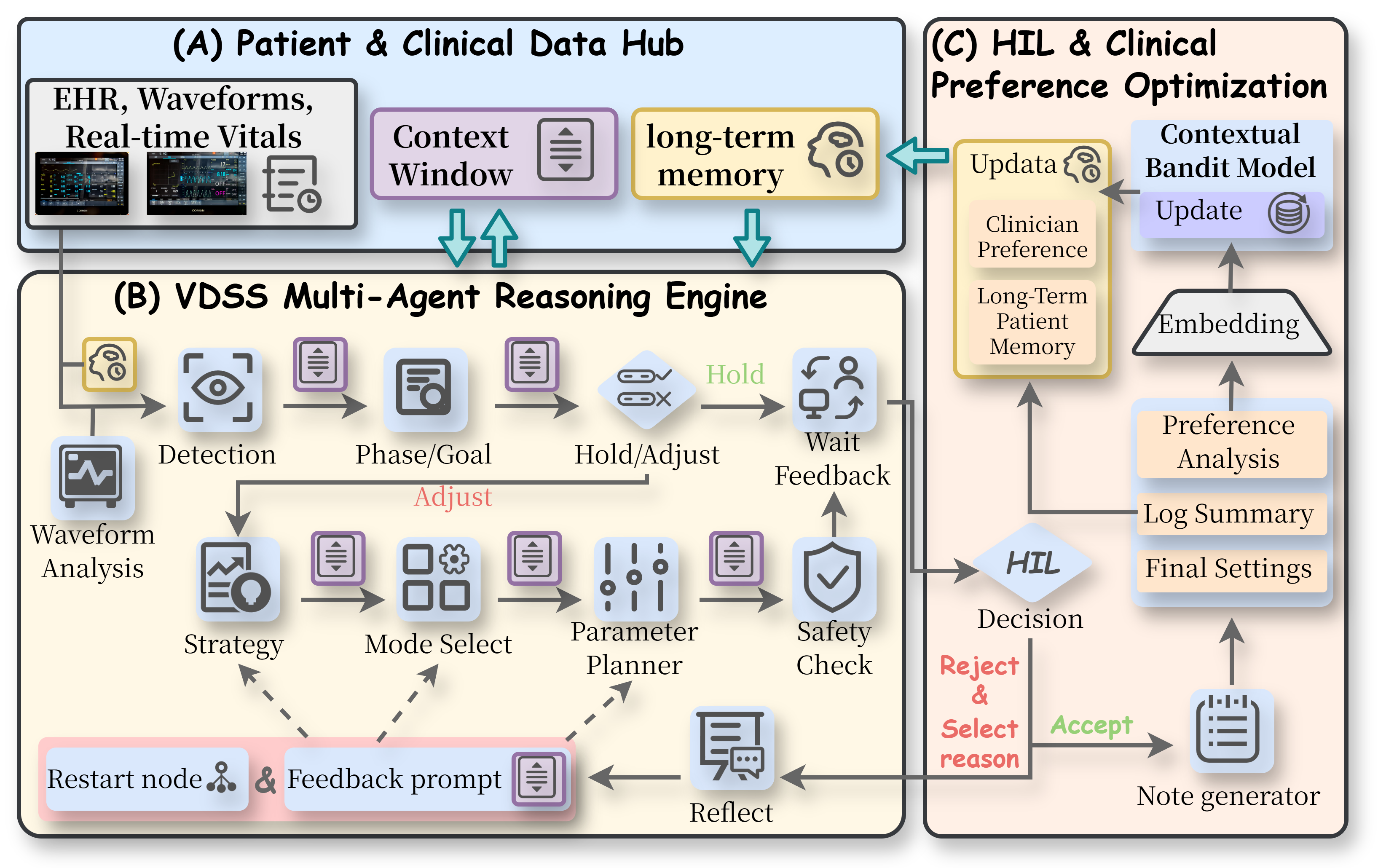}
  \caption{Overview of VDSS. Panel \textbf{A} shows the patient and clinical data hub with a context window and long term memory. Panel \textbf{B} depicts the VDSS multi agent reasoning engine from detection and phase goal inference to hold adjust gating, planning, safety checking, and feedback driven revision. Panel \textbf{C} summarizes the human in the loop interface and online clinician preference optimization with a contextual bandit.}
  \label{fig:vdss_overview}
\end{figure}

\section{Ventilator Decision Support System}
\subsection{VDSS Architecture: Multi-Agent Framework}

VDSS organizes bedside ventilator titration as a contract governed multi-agent workflow rather than a single model that directly maps clinical inputs to ventilator recommendations. Here, agents refer to LLM or VLM reasoning modules with defined clinical roles, role specific inputs, and schema constrained outputs, while graph routing, mode compatibility validation, and safety checking are implemented as deterministic control components. At each adjustment cycle $t$, VDSS forms the decision input $x_t$ from the current bedside state, current ventilator settings, short term context, and long term memory. Physiological and ventilatory measurements are treated as state variables, whereas clinician adjustable ventilator settings, including ventilation mode, PEEP, oxygen concentration, pressure support, inspiratory pressure, and respiratory rate, are treated as action variables. This design maintains a traceable path from evidence extraction to decision reasoning and clinician facing communication.

When waveform data are available, the \textit{\textbf{Waveform Analyzer}} first converts rendered pressure and flow traces into structured bedside cues, including waveform quality, asynchrony patterns, suspicious events, observed patient state, and uncertainty, thereby complementing tabular signals with dynamic ventilatory evidence \cite{thille2006patient}. The \textit{\textbf{Detection Agent}} integrates these waveform-derived cues with physiological and ventilatory measurements to identify abnormalities and generate a structured state summary. This summary is then interpreted by the \textit{\textbf{Phase Goal Manager}}, which infers the current treatment phase and specifies primary and secondary goals. Conditioned on the detected abnormalities and phase-consistent objectives, the \textit{\textbf{Hold/Adjust Gate}} selects a branch decision $b_t \in \{\mathrm{hold}, \mathrm{adjust}\}$. This gate enables VDSS to avoid unnecessary titration when the current configuration is acceptable or when the available evidence is insufficient for safe adjustment.

The overall workflow is illustrated in Fig.~\ref{fig:vdss_overview}. If $b_t=\mathrm{hold}$, VDSS proceeds directly to cycle closure by generating a clinician-facing summary and structured record. If $b_t=\mathrm{adjust}$, VDSS enters the active titration branch. The \textit{\textbf{Strategy Selector}} determines the adjustment priority, the \textit{\textbf{Mode Select}} agent evaluates whether a mode change is required under device- and mode-specific control semantics, and the \textit{\textbf{Parameter Planner}} proposes a compact set of executable setting updates over allowable action variables. Before clinician review, proposed updates are passed through deterministic safety and compatibility checks, ensuring that generative planning remains bounded by explicit device, mode, and safety constraints. If the clinician rejects a recommendation, the \textit{\textbf{Reflect Agent}} translates structured feedback into revision constraints and identifies the minimal decision layer to revisit, such as strategy selection, mode selection, or parameter planning. Once a recommendation is accepted, the \textit{\textbf{Note Generator}} produces the final clinician-facing summary, stores an auditable cycle record, and extracts the preference signal used for subsequent preference adaptation.

\subsection{Human in the Loop Interaction and Replanning}
VDSS is structured around clinician confirmed adjustment cycles. After the reasoning engine produces candidate setting updates and they pass an initial \textit{Safety} consistency check, the system pauses and presents the proposed ventilator mode and parameter updates together with the safety assessment and the current preference context. This cycle level checkpoint matches bedside titration practice and makes clinician oversight an explicit component of the control loop \cite{vasey2022reporting}.

For cycle $t$, interaction proceeds through rounds $k=1,\dots,K_t$. The proposal at round $k$ is $a_t^{(k)}$, and clinician feedback is $o_t^{(k)}\in\{\mathrm{accept},\mathrm{reject}\}$. The cycle ends at the first accepted proposal, yielding the final accepted configuration $\tilde a_t$:
\begin{equation}
K_t=\min\{k:\,o_t^{(k)}=\mathrm{accept}\}, \qquad \tilde a_t=a_t^{(K_t)} .
\end{equation}

Rejection is treated as a structured control signal. When a proposal is rejected, the system records the clinician rationale and the specific parameters at issue, and the \textit{Reflect Agent} translates this feedback into revision constraints while localizing the minimal decision level to revisit. Replanning is then applied only to the implicated stage, with intermediate results reused whenever possible rather than rerunning the full pipeline. When waveform input is available, the \textit{Waveform Analyzer} may be re executed to refresh evidence prior to revision so subsequent proposals remain grounded in current bedside cues. This targeted rollback mechanism keeps updates confined to the feedback relevant layer and reduces redundant inference across rounds.

Upon acceptance, the \textit{Note Generator} closes the cycle by producing a structured summary of the final decision and extracting a clinician preference signal from the completed interaction. These outputs are written to long term memory to support trajectory continuity and subsequent preference adaptation, while the within cycle trace is retained for audit and for the cycle end preference update described next. Interaction is bounded by a maximum round budget $K_{\max}$ to maintain a controlled and reviewable process.

\begin{algorithm}[t]
\caption{Cycle end preference update with a contextual bandit in VDSS}
\label{alg:vdss_bandit}
\begin{algorithmic}[1]
\State \textbf{Input:} clinician $d$, context $x_t$, current settings, long term memory
\State $k\gets 1$, $\mathrm{accepted}\gets \mathrm{false}$
\While{$\neg \mathrm{accepted}$ and $k\le K_{\max}$}
  \State Generate proposal $a_t^{(k)}$, apply \textit{Safety}, present for clinician review, collect $o_t^{(k)}$
  \If{$o_t^{(k)}=\mathrm{reject}$}
    \State \textit{Reflect Agent} produces revision constraints and selects the minimal decision stage to revisit
    \State Replan from the selected stage; $k\gets k+1$
  \Else
    \State $\tilde a_t\gets a_t^{(k)}$, $K_t\gets k$, $\mathrm{accepted}\gets \mathrm{true}$
  \EndIf
\EndWhile
\State \textit{Note Generator} outputs log summary $s_t$ and preference signal $p_t$
\State Update clinician preference state $\theta_d \leftarrow \mathrm{BanditUpdate}(\theta_d;\,x_t,\tilde a_t,h_t,p_t)$
\State Write $s_t$ and $p_t$ to long term memory
\end{algorithmic}
\end{algorithm}

\subsection{Online Preference Optimization with Contextual Bandit}
VDSS adapts to clinician specific tuning styles via an online contextual bandit that is updated only after an adjustment cycle is resolved by clinician acceptance. For each clinician $d$, the bandit maintains a preference state $\theta_d$ \cite{lei2017actor}. This state is provided to the reasoning engine as preference context and modulates candidate ranking and selection in subsequent cycles. At cycle closure, \textit{Note Generator} produces two cycle level artifacts that are persisted in long term memory, namely a final patient log summary and a clinician preference signal.

Preference learning is defined over a fixed set of twelve long term preference categories, which correspond to the twelve bandit arms used for clinician modeling. These categories span preferences over adjustment granularity and adjustment style, as well as emphasis across clinical objectives and pragmatic workflow choices commonly observed in bedside titration. Specifically, the categories include preferences for mode level changes versus staying within the current mode, conservative small step updates versus target driven assertive adjustments, and prioritization of oxygenation, ventilation and acid base control, lung protection, hemodynamics, synchrony and comfort, and weaning. They further encode a tendency to focus on a single key parameter first and a tendency to defer action when available data or context are insufficient.

Let $x_t$ denote the cycle context assembled from the data hub and memory, and let
$h_t=\{(a_t^{(k)},o_t^{(k)})\}_{k=1}^{K_t}$
denote the full interaction trace for cycle $t$. After a proposal is accepted, the final accepted settings $\tilde a_t$ and the preference signal $p_t$ produced at cycle closure define a single cycle end update:
\begin{equation}
\theta_d \leftarrow \mathrm{BanditUpdate}\!\left(\theta_d;\,x_t, \tilde a_t, h_t, p_t\right).
\end{equation}
This update is driven by the clinician approved outcome and the full cycle trace rather than intermediate rejected proposals, aligning preference adaptation with bedside decision making.

\section{Experiments}
\label{sec:experiments}

We evaluate VDSS using retrospective ICU trajectory replay with cycle based clinician review. The dataset is collected from a multi center ICU cohort and includes 1309 structured records and 7447 ventilator setting entries covering 13 ventilation modes across two ventilator brands, with the four most frequent modes accounting for 83.4\% of all entries. We quantify next step replay by predicting the next clinically recorded setting update from the current bedside state and settings, reporting normalized error metrics and averaged $R^2$.

Table~\ref{tab:main_results} summarizes performance across replay and expert review. Across backbones, deploying models within VDSS yields substantially better replay accuracy and clinician ratings than direct single model generation. For example, under GPT-5.2, VDSS reduces MSE from 0.343 to 0.102 while improving the overall rating from 2.63 to 4.11. The ablations show consistent degradation when removing waveform evidence or preference context, with MSE increasing to 0.145 (NoImg) or 0.128 (NoPref) and overall ratings dropping to 3.62 and 3.89, respectively, indicating that both components contribute to more stable recommendations and clearer clinician facing justification. Similar gains are observed across other backbones, suggesting that the improvement is driven by the VDSS workflow rather than any single model.

\begin{table}[t]
\centering
\caption{\textbf{Main results.} Replay metrics are computed on the full dataset and clinician ratings (1--5) are from expert review of 100 cycles for single model baselines, VDSS, and VDSS ablations.}
\label{tab:main_results}
\begin{tabular}{l c c c c c c c}
\toprule
& \multicolumn{3}{c}{\textbf{Replay metrics}} & \multicolumn{4}{c}{\textbf{Clinician ratings (1--5)}} \\
\cmidrule(lr){2-4}\cmidrule(lr){5-8}
\textbf{Method} &
\textbf{MSE} $\downarrow$ &
\textbf{MAE} $\downarrow$ &
\textbf{$R^2$} $\uparrow$ &
\textbf{Accept.} &
\textbf{Safety} &
\textbf{Clarity} &
\textbf{Overall} \\
\midrule
\multicolumn{8}{l}{\textbf{Single model end to end}} \\
Qwen3-VL-8B           & 0.486 & 0.595 & -0.132 & 2.45 & 2.11 & 2.26 & 2.27 \\
Qwen3-VL-235B         & 0.390 & 0.426 & 0.019  & 2.53 & 2.17 & 2.19 & 2.30 \\
Gemini 3 Pro          & 0.362 & 0.386 & 0.130  & 2.80 & 2.31 & 2.24 & 2.45 \\
GPT-5.2               & 0.343 & 0.381 & 0.153  & 2.75 & 2.43 & 2.71 & 2.63 \\
\midrule
\multicolumn{8}{l}{\textbf{VDSS (same backbone, multi agent)}} \\
Qwen3(VL)-8B (VDSS)   & 0.175 & 0.210 & 0.354  & 3.30 & 4.12 & 3.56 & 3.22 \\
Qwen3-VL-235B (VDSS)  & 0.130 & 0.183 & 0.661  & 3.51 & 4.20 & 3.79 & 3.59 \\
Gemini 3 Pro (VDSS)   & 0.124 & 0.166 & 0.693  & 3.80 & 4.30 & 3.91 & 3.82 \\
GPT-5.2 (VDSS)        & \textbf{0.102} & \textbf{0.162} & \textbf{0.743} & \textbf{4.09} & \textbf{4.46} & \textbf{4.15} & \textbf{4.11} \\
\midrule
\multicolumn{8}{l}{\textbf{Ablations (GPT-5.2 within VDSS)}} \\
NoImg                & 0.145 & 0.194 & 0.660  & 3.69 & 4.26 & 3.75 & 3.62 \\
NoPref               & 0.128 & 0.183 & 0.707  & 3.81 & 4.32 & 3.95 & 3.89 \\
NoImgNoPref          & 0.158 & 0.202 & 0.620  & 3.50 & 4.15 & 3.61 & 3.46 \\
\bottomrule
\end{tabular}
\end{table}

\begin{figure}[t]
  \centering
  \includegraphics[width=\linewidth]{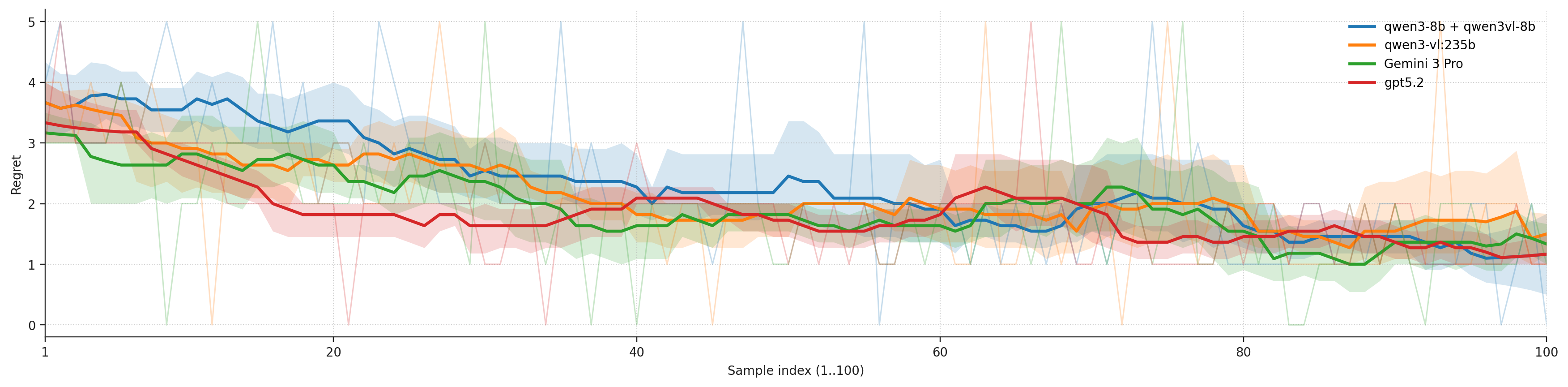}
  \caption{Regret over 100 cycles from a single clinician, defined as $K_t-1$ and capped by $K_{\max}$, shows an overall downward trend.}
  \label{fig:regret_line}
\end{figure}

We do not emphasize conventional reinforcement learning because ventilation modes differ substantially in parameter sets and semantics, and per mode data are comparatively limited, which degrades generalization \cite{liu2020reinforcement,peine2021development}. The high risk setting further motivates a reviewable interaction loop with explicit safety checks and an auditable evidence trail, which is difficult to guarantee with end to end policies across heterogeneous modes.

We assess interaction efficiency through cycle level rejection dynamics on the same 100 cycle study. For each cycle $t$, interaction proceeds across rounds $k=1,\dots,K_t$ and ends at the first accepted proposal. Regret within a cycle is the number of rejected proposals, $K_t-1$, capped by the maximum round budget $K_{\max}$. Fig.~\ref{fig:regret_line} shows an overall downward regret trend across model variants, consistent with improved interaction efficiency as preference context is updated.

We report runtime on an RTX 5070 Ti for local inference with Qwen3-8B and Qwen3-VL-8B. The workflow averages 305.3 s per cycle, and waveform analysis is the main bottleneck at 56.46 s. With external APIs, latency can vary with network conditions and server load. The framework maintains a completion failure rate below \textbf{7\%}.

\section{Case Study}
\label{sec:case_study}

\begin{figure}[t]
  \centering
  \includegraphics[width=\linewidth]{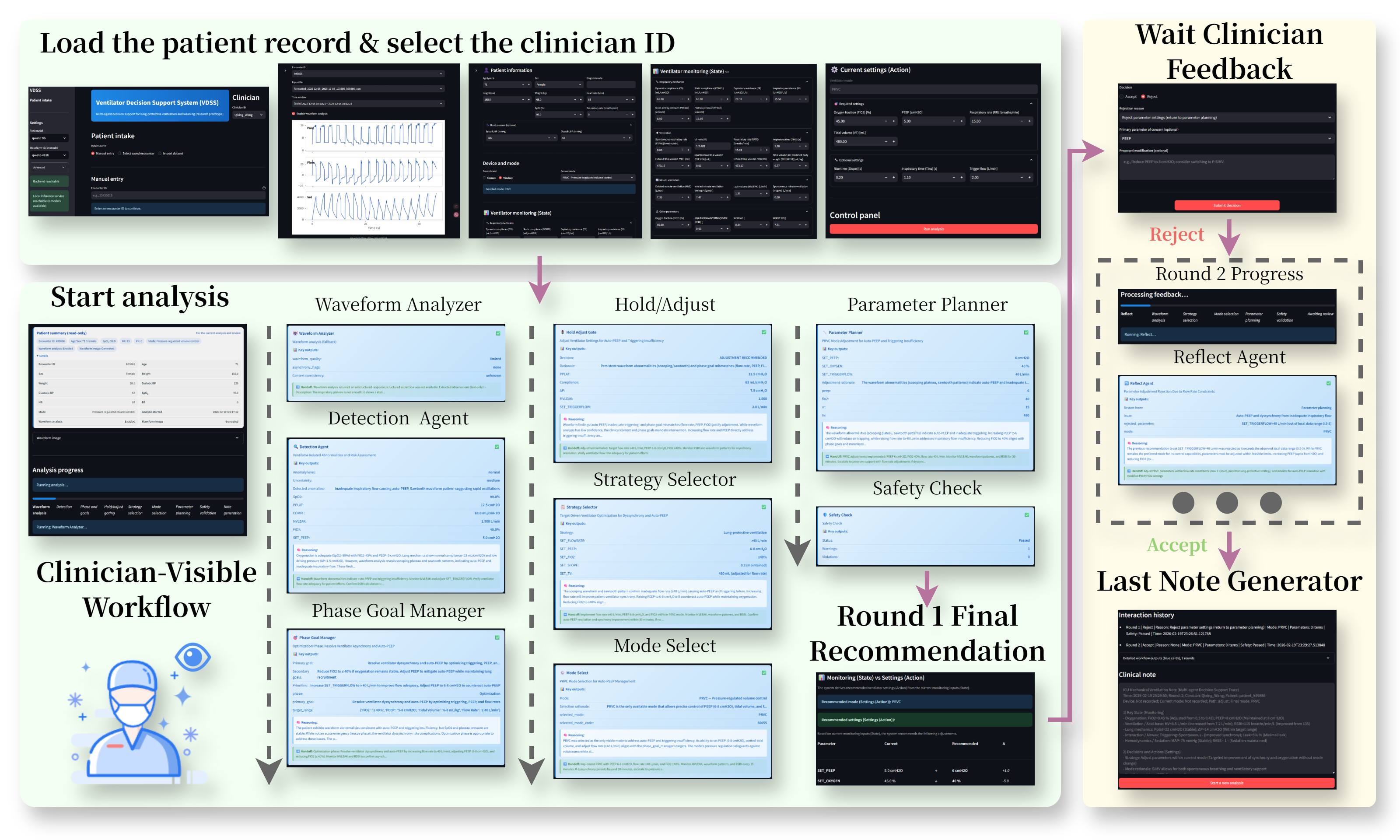}
  \caption{Case study of one VDSS adjustment cycle.}
  \label{fig:case_study_workflow}
\end{figure}
Fig.~\ref{fig:case_study_workflow} illustrates a complete adjustment cycle in the prototype interface. The clinician selects a clinician ID, loads a patient encounter, specifies the analysis time window, and enables waveform analysis when available. The patient is managed in PRVC with stable oxygenation and lung mechanics, while pressure and flow waveforms show a scooped inspiratory plateau and sawtooth patterns consistent with insufficient inspiratory flow and elevated auto PEEP risk. The system summarizes abnormalities and evidence, defines phase consistent objectives to improve synchrony and reduce auto PEEP risk while de escalating FiO$2$ when appropriate, and maintains PRVC while proposing a small set of safety checked parameter updates for review. The first proposal is rejected due to local feasibility constraints, which are converted into explicit revision constraints to replan from the implicated stage, yielding an updated recommendation accepted in the second round. The cycle ends with a structured note and an auditable interaction log written to long term memory, and the accepted outcome updates clinician specific preferences for subsequent cycles.

\section{Conclusion}
\label{sec:conclusion}
We present a human in the loop multi agent ventilator decision support framework that combines contract governed modules with traceable evidence and safety checked recommendations, enabling interpretable and reviewable bedside titration. Retrospective ICU trajectory replay and expert review show improved acceptability and fewer interaction rounds, indicating that the framework is effective in practice. An online contextual bandit updates clinician specific preferences at cycle closure, which improves alignment with individual tuning styles and supports consistent, preference aware recommendations under the same safety boundaries.

\begin{credits}
\subsubsection{\ackname}
This work was supported by the Shanghai Municipal Science and Technology Commission (Grant No. 25692114700).

\subsubsection{\discintname}
The authors have no competing interests to declare that are relevant to the content of this article.
\end{credits}
%
%
%
%
%
%
%
%
\bibliographystyle{splncs04}
\bibliography{refs}

@article{slutsky2013ventilator,
  title={Ventilator-induced lung injury},
  author={Slutsky, Arthur S and Ranieri, V Marco},
  journal={New England Journal of Medicine},
  volume={369},
  number={22},
  pages={2126--2136},
  year={2013},
  publisher={Mass Medical Soc}
}

@article{topol2019high,
  title={High-performance medicine: the convergence of human and artificial intelligence},
  author={Topol, Eric J},
  journal={Nature medicine},
  volume={25},
  number={1},
  pages={44--56},
  year={2019},
  publisher={Nature Publishing Group US New York}
}

@article{murali2024clinical,
  title={Clinical practice, decision-making, and use of clinical decision support systems in invasive mechanical ventilation: a narrative review},
  author={Murali, Mayur and Ni, Melody and Karbing, Dan S and Rees, Stephen E and Komorowski, Matthieu and Marshall, Dominic and Ramnarayan, Padmanabhan and Patel, Brijesh V},
  journal={British Journal of Anaesthesia},
  volume={133},
  number={1},
  pages={164--177},
  year={2024},
  publisher={Elsevier}
}

@article{acute2000ventilation,
  title={Ventilation with lower tidal volumes as compared with traditional tidal volumes for acute lung injury and the acute respiratory distress syndrome},
  author={Acute Respiratory Distress Syndrome Network},
  journal={New England Journal of Medicine},
  volume={342},
  number={18},
  pages={1301--1308},
  year={2000},
  publisher={Mass Medical Soc}
}

@article{jouvet2011development,
  title={Development and implementation of explicit computerized protocols for mechanical ventilation in children},
  author={Jouvet, Philippe and Hernert, Patrice and Wysocki, Marc},
  journal={Annals of intensive care},
  volume={1},
  number={1},
  pages={51},
  year={2011},
  publisher={Elsevier}
}

@inproceedings{wu2024autogen,
  title={Autogen: Enabling next-gen LLM applications via multi-agent conversations},
  author={Wu, Qingyun and Bansal, Gagan and Zhang, Jieyu and Wu, Yiran and Li, Beibin and Zhu, Erkang and Jiang, Li and Zhang, Xiaoyun and Zhang, Shaokun and Liu, Jiale and others},
  booktitle={First conference on language modeling},
  year={2024}
}

@inproceedings{dewes2025contract,
  title={Contract-based design and verification of multi-agent systems with quantitative temporal requirements},
  author={Dewes, Rafael and Dimitrova, Rayna},
  booktitle={Proceedings of the AAAI Conference on Artificial Intelligence},
  volume={39},
  number={22},
  pages={23152--23159},
  year={2025}
}

@article{packer2023memgpt,
  title={MemGPT: towards LLMs as operating systems.},
  author={Packer, Charles and Fang, Vivian and Patil, Shishir\_G and Lin, Kevin and Wooders, Sarah and Gonzalez, Joseph\_E},
  year={2023},
  publisher={ArXiv}
}

@article{vasey2022reporting,
  title={Reporting guideline for the early stage clinical evaluation of decision support systems driven by artificial intelligence: DECIDE-AI},
  author={Vasey, Baptiste and Nagendran, Myura and Campbell, Bruce and Clifton, David A and Collins, Gary S and Denaxas, Spiros and Denniston, Alastair K and Faes, Livia and Geerts, Bart and Ibrahim, Mudathir and others},
  journal={bmj},
  volume={377},
  year={2022},
  publisher={British Medical Journal Publishing Group}
}

@article{liu2020reinforcement,
  title={Reinforcement learning for clinical decision support in critical care: comprehensive review},
  author={Liu, Siqi and See, Kay Choong and Ngiam, Kee Yuan and Celi, Leo Anthony and Sun, Xingzhi and Feng, Mengling},
  journal={Journal of medical Internet research},
  volume={22},
  number={7},
  pages={e18477},
  year={2020},
  publisher={JMIR Publications Toronto, Canada}
}

@article{peine2021development,
  title={Development and validation of a reinforcement learning algorithm to dynamically optimize mechanical ventilation in critical care},
  author={Peine, Arne and Hallawa, Ahmed and Bickenbach, Johannes and Dartmann, Guido and Fazlic, Lejla Begic and Schmeink, Anke and Ascheid, Gerd and Thiemermann, Christoph and Schuppert, Andreas and Kindle, Ryan and others},
  journal={NPJ digital medicine},
  volume={4},
  number={1},
  pages={32},
  year={2021},
  publisher={Nature Publishing Group UK London}
}

@article{fan2017official,
  title={An official American Thoracic Society/European Society of Intensive Care Medicine/Society of Critical Care Medicine clinical practice guideline: mechanical ventilation in adult patients with acute respiratory distress syndrome},
  author={Fan, Eddy and Del Sorbo, Lorenzo and Goligher, Ewan C and Hodgson, Carol L and Munshi, Laveena and Walkey, Allan J and Adhikari, Neill KJ and Amato, Marcelo BP and Branson, Richard and Brower, Roy G and others},
  journal={American journal of respiratory and critical care medicine},
  volume={195},
  number={9},
  pages={1253--1263},
  year={2017},
  publisher={American Thoracic Society}
}

@article{goodfellow2024aarc,
  title={AARC clinical practice guideline: patient-ventilator assessment},
  author={Goodfellow, Lynda T and Miller, Andrew G and Varekojis, Sarah M and LaVita, Carolyn J and Glogowski, Joel T and Hess, Dean R},
  journal={Respiratory Care},
  volume={69},
  number={8},
  pages={1042--1054},
  year={2024},
  publisher={SAGE Publications Sage CA: Los Angeles, CA}
}

@article{gray2021changes,
  title={Changes in stress and workplace shortages reported by US critical care physicians treating coronavirus disease 2019 patients},
  author={Gray, Bradley M and Vandergrift, Jonathan L and Barnhart, Brendan J and Reddy, Siddharta G and Chesluk, Benjamin J and Stevens, Jennifer S and Lipner, Rebecca S and Lynn, Lorna A and Barnett, Michael L and Landon, Bruce E},
  journal={Critical Care Medicine},
  volume={49},
  number={7},
  pages={1068--1082},
  year={2021},
  publisher={LWW}
}

@article{ranieri2012acute,
  title={Acute respiratory distress syndrome: the Berlin Definition.},
  author={Ranieri, V Marco and Rubenfeld, Gordon D and Taylor Thompson, B and Ferguson, Niall D and Caldwell, Ellen and Fan, Eddy and Camporota, Luigi and Slutsky, Arthur S},
  journal={JAMA: Journal of the American Medical Association},
  volume={307},
  number={23},
  year={2012}
}

@article{liu2020reporting,
  title={Reporting guidelines for clinical trial reports for interventions involving artificial intelligence: the CONSORT-AI extension},
  author={Liu, Xiaoxuan and Rivera, Samantha Cruz and Moher, David and Calvert, Melanie J and Denniston, Alastair K and Ashrafian, Hutan and Beam, Andrew L and Chan, An-Wen and Collins, Gary S and Deeks, Ara DarziJonathan J and others},
  journal={The Lancet Digital Health},
  volume={2},
  number={10},
  pages={e537--e548},
  year={2020},
  publisher={Elsevier}
}

@article{thille2006patient,
  title={Patient-ventilator asynchrony during assisted mechanical ventilation},
  author={Thille, Arnaud W and Rodriguez, Pablo and Cabello, Belen and Lellouche, Fran{\c{c}}ois and Brochard, Laurent},
  journal={Intensive care medicine},
  volume={32},
  number={10},
  pages={1515--1522},
  year={2006},
  publisher={Springer}
}

@article{lei2017actor,
  title={An actor-critic contextual bandit algorithm for personalized mobile health interventions},
  author={Lei, Huitian and Lu, Yangyi and Tewari, Ambuj and Murphy, Susan A},
  journal={arXiv preprint arXiv:1706.09090},
  year={2017}
}

@article{shi2023nurses,
  title={How do nurses manage their work under time pressure? Occurrence of implicit rationing of nursing care in the intensive care unit: A qualitative study},
  author={Shi, Fang and Li, Yuntao and Zhao, Yingnan},
  journal={Intensive and Critical Care Nursing},
  volume={75},
  pages={103367},
  year={2023},
  publisher={Elsevier}
}

@article{henry2022human,
  title={Human--machine teaming is key to AI adoption: clinicians’ experiences with a deployed machine learning system},
  author={Henry, Katharine E and Kornfield, Rachel and Sridharan, Anirudh and Linton, Robert C and Groh, Catherine and Wang, Tony and Wu, Albert and Mutlu, Bilge and Saria, Suchi},
  journal={NPJ digital medicine},
  volume={5},
  number={1},
  pages={97},
  year={2022},
  publisher={Nature Publishing Group UK London}
}

@article{jivraj2023use,
  title={Use of mechanical ventilation across 3 countries},
  author={Jivraj, Naheed K and Hill, Andrea D and Shieh, Meng-Shiou and Hua, May and Gershengorn, Hayley B and Ferrando-Vivas, Paloma and Harrison, David and Rowan, Kathy and Lindenauer, Peter K and Wunsch, Hannah},
  journal={JAMA internal medicine},
  volume={183},
  number={8},
  pages={824--831},
  year={2023}
}

@article{von2024adjustments,
  title={Adjustments of ventilator parameters during operating room--to--ICU transition and 28-day mortality},
  author={von Wedel, Dario and Redaelli, Simone and Suleiman, Aiman and Wachtendorf, Luca J and Fosset, Maxime and Santer, Peter and Shay, Denys and Munoz-Acuna, Ricardo and Chen, Guanqing and Talmor, Daniel and others},
  journal={American Journal of Respiratory and Critical Care Medicine},
  volume={209},
  number={5},
  pages={553--562},
  year={2024},
  publisher={Oxford University Press}
}

@article{rosenbacke2024explainable,
  title={How explainable artificial intelligence can increase or decrease clinicians’ trust in AI applications in health care: systematic review},
  author={Rosenbacke, Rikard and Melhus, {\AA}sa and McKee, Martin and Stuckler, David},
  journal={Jmir Ai},
  volume={3},
  pages={e53207},
  year={2024},
  publisher={JMIR Publications Toronto, Canada}
}

@article{munoz2025artificial,
  title={Artificial intelligence in the management of patient-ventilator asynchronies: A scoping review},
  author={Mu{\~n}oz, Javier and Ru{\'\i}z-Cacho, Roc{\'\i}o and Fern{\'a}ndez-Araujo, Nerio Jos{\'e} and Candela, Alberto and Visedo, Lourdes Carmen and Mu{\~n}oz-Visedo, Javier},
  journal={Heart \& Lung},
  volume={73},
  pages={139--152},
  year={2025},
  publisher={Elsevier}
}

@article{bion2010human,
  title={Human factors in the management of the critically ill patient},
  author={Bion, Jean Francis and Abrusci, T and Hibbert, P},
  journal={British journal of anaesthesia},
  volume={105},
  number={1},
  pages={26--33},
  year={2010},
  publisher={Oxford University Press}
}

@article{cox2019effects,
  title={Effects of a personalized web-based decision aid for surrogate decision makers of patients with prolonged mechanical ventilation: a randomized clinical trial},
  author={Cox, Christopher E and White, Douglas B and Hough, Catherine L and Jones, Derek M and Kahn, Jeremy M and Olsen, Maren K and Lewis, Carmen L and Hanson, Laura C and Carson, Shannon S},
  journal={Annals of internal medicine},
  volume={170},
  number={5},
  pages={285--297},
  year={2019},
  publisher={American College of Physicians}
}

@article{kollef2000effect,
  title={The effect of respiratory therapist-initiated treatment protocols on patient outcomes and resource utilization},
  author={Kollef, Marin H and Shapiro, Steven D and Clinkscale, Darnetta and Cracchiolo, Lisa and Clayton, Donna and Wilner, Russ and Hossin, Linda},
  journal={Chest},
  volume={117},
  number={2},
  pages={467--475},
  year={2000},
  publisher={Elsevier}
}

\end{document}